%% file: main_cr.tex
\documentclass[10pt,twocolumn,letterpaper]{article}

\usepackage{cvpr}              %

\input{preamble}
\definecolor{cvprblue}{rgb}{0.21,0.49,0.74}
\usepackage{amssymb}
\usepackage{amsmath}
\usepackage{bbm}
\usepackage{gensymb}
\usepackage{multirow}
\usepackage{graphicx}
\usepackage[pagebackref,breaklinks,colorlinks,allcolors=cvprblue]{hyperref}
\def\paperID{2153} %
\def\confName{CVPR}
\def\confYear{2025}

\definecolor{commentgreen}{RGB}{2,200,10}

\title{FlexGS: Train Once, Deploy Everywhere with Many-in-One \\Flexible 3D Gaussian Splatting}

\author{Hengyu Liu$^{1}$\thanks{denotes Equal Contribution, $^\dagger$ denotes Project Lead.}, Yuehao Wang$^{2*}$, Chenxin Li$^{1*}$, Ruisi Cai$^{2, 3\dagger}$, Kevin Wang$^{2}$, \\
Wuyang Li$^{1}$, Pavlo Molchanov$^{3}$, Peihao Wang$^{2}$, Zhangyang Wang$^{2}$ \\
$^1$ The Chinese University of Hong Kong, $^2$ University of Texas at Austin, $^3$ Nvidia
}

\begin{document}

\maketitle
\input{sec/0_abstract}    
\input{sec/1_intro}
\input{sec/2_related}

\input{sec/3_method}

\input{sec/4_exp}
\input{sec/5_conclusion}
\input{}
\clearpage
{
    \small
    \bibliographystyle{ieeenat_fullname}
    \bibliography{main}
}

\appendix

\setcounter{page}{1}
\maketitlesupplementary

\section{Theoretical Analysis}
\label{sec:theory_analysis}
\subsection{Differentiable Gaussian Selection}
To substantiate the claims made in Sec. 3.2 regarding the ability of the proposed GsNet and Gumbel-Softmax~\cite{jang2016categorical} mechanisms to enable Adaptive Gaussian Selection in a differentiable manner, the following derivation is provided. 
As shown in Eq.~\ref{init-noise} and Eq.~\ref{add-noise}, with the logits $\boldsymbol{z} \in \mathbb{R}^{N\times C}$, where $N$ represent the number of Gaussians and $C$ denotes to the number of classes (set to $2$), a Gumbel noise sampling is conducted, whereby noise is integrated and the temperature parameter is appropriately scaled by $\tau$.
\begin{equation}
g_{i,c} = -\log\left(-\log\left(U_{i,c}\right)\right), U_{i,c} \sim \text{Uniform}(0,1).
\label{init-noise}
\end{equation}
\begin{equation}
\boldsymbol{\tilde{z}}_{i,c} = \frac{\boldsymbol{z}_{i,c} + g_{i,c}}{\tau}.
\label{add-noise}
\end{equation}
Then a softmax function is used for calculating soft output.
\begin{equation}
\boldsymbol{z}_{\text{soft},i,c} = \frac{\exp\left(\boldsymbol{\tilde{z}}_{i,c}\right)}{\sum_{k=1}^{C} \exp\left(\boldsymbol{\tilde{z}}_{i,k}\right)}
\label{softmax}
\end{equation}
Alternatively, discrete hard outputs may be derived from the soft outputs for utilization in forward propagation.
\begin{equation}
\label{hard-output}
\boldsymbol{z}_{\text{hard},i,c} =
\begin{cases}
1 & \text{if } c = \arg\max_{k} \boldsymbol{z}_{\text{soft},i,k} \\
0 & \text{otherwise}
\end{cases}
\end{equation}

The \textbf{Straight-Through Estimator} is employed to reconcile the discrete nature of hard outputs with the differentiable characteristics of soft outputs within the hard Gumbel-Softmax framework:
\begin{equation}
\begin{aligned}
B_{i} =& \boldsymbol{z}_{\text{hard},i} - \boldsymbol{z}_{\text{soft},i} + \boldsymbol{z}_{\text{soft},i} \\
=& \boldsymbol{z}_{\text{hard},i} + \left( \boldsymbol{z}_{\text{soft},i} - \boldsymbol{z}_{\text{soft},i} \right)  \\
=& \boldsymbol{z}_{\text{hard},i} + \text{stop\_gradient}(\boldsymbol{z}_{\text{soft},i})
\end{aligned}
\end{equation}
where \(\text{stop\_gradient}(\boldsymbol{z}_{\text{soft},i})\) signifies that during backpropagation, the gradient associated with \( \boldsymbol{z}_{\text{soft},i} \) is disregarded, thereby exclusively preserving the value of \( \boldsymbol{z}_{\text{hard},i} \).

For an entire batch of size $N$, let the input matrix $\boldsymbol{z} \in \mathbb{R}^{N \times 2}$ and the output matrix $B \in \mathbb{R}^{N \times 2}$ be defined. While the selected mask $\boldsymbol{\hat{M}} = \boldsymbol{z}_{output}[:,\, -1]$. The gradient matrix $\frac{\partial \boldsymbol{\hat{M}}}{\partial \boldsymbol{z}} \in \mathbb{R}^{N \times 2}$ is delineated as follows:

\begin{equation}
\frac{\partial \boldsymbol{\hat{M}}}{\partial \boldsymbol{z}} =
\frac{1}{\tau}
\begin{bmatrix}
- B_{1,0} B_{1,1} & B_{1,1} \left(1 - B_{1,1}\right) \\
- B_{2,0} B_{2,1} & B_{2,1} \left(1 - B_{2,1}\right) \\
\vdots & \vdots \\
- B_{N,0} B_{N,1} & B_{N,1} \left(1 - B_{N,1}\right)
\end{bmatrix}
\end{equation}

where, the temperature parameter is denoted by \(\tau\). The probabilities of the \(i\)-th sample belonging to class \(0\) and class \(1\) are represented by \( B_{i0} \) and \( B_{i1} \), respectively. The ellipsis indicates that this pattern continues similarly for all $N$ samples. The aforementioned gradient matrix can also be expressed in a vectorized form as follows:
\begin{equation}
\frac{\partial \boldsymbol{\hat{M}}}{\partial \boldsymbol{z}} =
\frac{1}{\tau}
\begin{bmatrix}
- B[:, 0] \odot B[:,1]&B[:,1] \odot (1 - B[:, 1])
\end{bmatrix},
\label{gradient}
\end{equation}
where \( \odot \) signifies element-wise multiplication. From Eq.~\ref{gradient}, we can observe that the gradient of each parameter in GsNet can be calculated based on the ``chain rule".

\subsection{Gradients of Gaussian Attributes}
To elucidate the computational procedure, we hereby redefine the notations previously employed in Sec. 3.1. The current opacity of the specific Gaussian $i$ within the rendering process for pixel $p$ is illustrated in Eq.~\ref{render-ith}.
\begin{equation}
\alpha_i = \hat{o}_i \cdot  \boldsymbol{G}_2(i, p),\quad \hat{o}_i= {o}_i \cdot \boldsymbol{\hat{M}}_i,
\label{render-ith}
\end{equation}
where $\hat{o}_i$ is the masked opacity for the selected Gaussians and $\boldsymbol{G}_2(i, p)$ denotes the effect coefficient of the 2D projection of the Gaussian $i$ to the pixel $p$.

With the given ratio $e$, the rendering loss of the selected Gaussian can be calculated as below:
\begin{equation}
\mathcal{L}_{s}=|\boldsymbol{I}_s^{e}-\boldsymbol{I}_{GT}|.
\end{equation}
Specially, for the $i-$th Gaussian interacted with pixel $p$ on rendered Image $\boldsymbol{I}_s^{e}$, the gradient of the Gaussian attribute $\mu$ can be calculated as shown in Eq.~\ref{gradient-mu}.

\begin{equation}
\begin{aligned}
\frac{\partial \mathcal{L}_{s}}{\partial \mu_i}=& \frac{\partial \mathcal{L}_{s}}{\partial \boldsymbol{\hat{o}}_i}\cdot \frac{\partial \boldsymbol{\hat{o}}_i}{\partial \mu_i}\cdot\boldsymbol{G}_2 + \frac{\partial \mathcal{L}_{s}}{\partial\boldsymbol{G}_2} \cdot \frac{\partial\boldsymbol{G}_2}{\mu_i}\cdot \boldsymbol{\hat{o}}_i \\
=&
\boldsymbol{G}_2 \cdot
\frac{\partial \mathcal{L}_{s}}{\partial \boldsymbol{\hat{o}}_i} 
\frac{\partial {o_i \cdot \boldsymbol{\hat{M}}_i}}{\partial \mu_i}
 +
o_i\boldsymbol{\hat{M}}_i
\cdot
\frac{\partial \mathcal{L}_{s}}{\partial\boldsymbol{G}_2}
\frac{\partial\boldsymbol{G}_2}{\mu_i}
\\
=&
o_i\boldsymbol{G}_2 \cdot 
\frac{\partial \mathcal{L}_{s}}{\partial \boldsymbol{\hat{o}}_i}
{\frac{\partial \boldsymbol{\hat{M}}_i}{\partial \boldsymbol{z}}}
\frac{\partial \boldsymbol{z}}{\partial \mu_i} 
+o_i\boldsymbol{\hat{M}}_i\cdot
\frac{\partial \mathcal{L}_{s}}{\partial\boldsymbol{G}_2}
\frac{\partial\boldsymbol{G}_2}{\mu_i}
\end{aligned}
\label{gradient-mu}
\end{equation}
Where $\frac{\partial \boldsymbol{z}}{\partial \mu_i}$ is the gradient of GsNet to $\mu_i$. Other attributes of Gaussians can also be calculated in the same process, with the gradient ${\frac{\partial \boldsymbol{\hat{M}}_i}{\partial \boldsymbol{z}}}$ is calculated in ~\ref{gradient}.
\begin{figure*}[t]
\begin{center}
  \includegraphics[trim={7 0 7 0},clip,width=\textwidth]{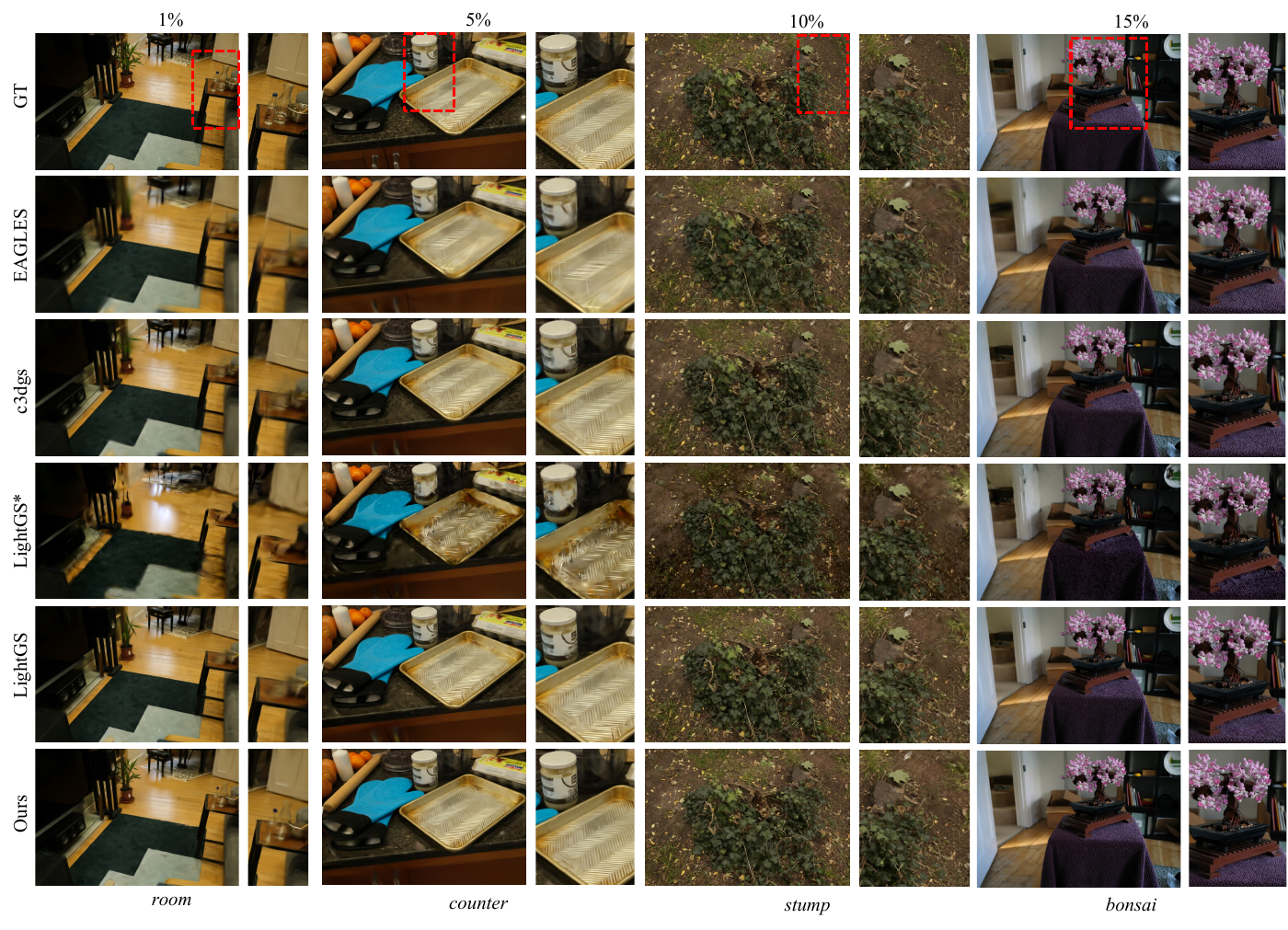}
  
\caption{
Visual results compared with other methods on various elastic ratios:\{0.01,0.05,0.10,0.15\} on Mip-NeRF360~\cite{mip360}:~\{\textit{bicycle}, \textit{room, counter, stump, bonsai}\}.
}
\label{fig:main-extra}
\end{center}
\end{figure*}

\section{Implementation Details}
\label{sec:more_details}
\subsection{Training Details}
In Sec. 3.2, the dimensionality $D$ of the hidden features in GsNet, employed for adaptive selection, is set to 64. To enhance computational efficiency, the implementation of the Spatial-Ratio Neural Field proposed in Sec. 3.3 adheres to the configurations outlined in~\cite{fridovich2023k,wu20244d}, utilizing six planes $\{(x,y), (x,z), (y,z), (x,e), (y,e), (z,e)\}$ to model the 4D voxel space. The resolutions across the four dimensions ($x$, $y$, $z$, and $e$) are configured as $\{64,\, 64,\, 64,\, 100\}$. Additionally, the hidden feature dimension of the Multi-Head Predictor for predicting the transformation under the given ratio is set to $64$.
\subsection{Inference Details}
During elastic inference, in contrast to the training phase where the opacity of Gaussians is multiplied by the binary mask values, we directly discard the unselected Gaussians. Furthermore, we observe that despite enforcing sparsity supervision on the masks predicted by GsNet, the number of activated entries within the predicted mask does not exactly achieve the desired ratio. For instance, a target ratio of $0.20$ results in approximately selecting $19.5\%$ of all Gaussians. Therefore, to attain an accurate elastic ratio, during inference, we employ Pytorch's \texttt{F.gumbel\_softmax} function with its parameter \texttt{hard=False} to output continuous logits and select the top $\lfloor eN \rfloor$ logits out of $N$.

\section{More Experimental Results}
\label{sec:exp_details}
In this section, we provide more pre-scene results. Visual comparisons on four scenes of Mip-NeRF360~\cite{mip360} \{\textit{bonsai}, \textit{counter}, \textit{room}, \textit{stump}\} under the given ratios $\{0.01, 0.05, 0.10, 0.15\}$ are shown in Fig.~\ref{fig:main-extra}. The breakdown results of quantitative comparisons on each scene of the tested datasets are from Tab.~\ref{mipnerf-bicycle} to Tab.~\ref{zip-london}.

\begin{figure*}[t]
\begin{center}
  \includegraphics[trim={7 0 7 0},clip,width=\textwidth]{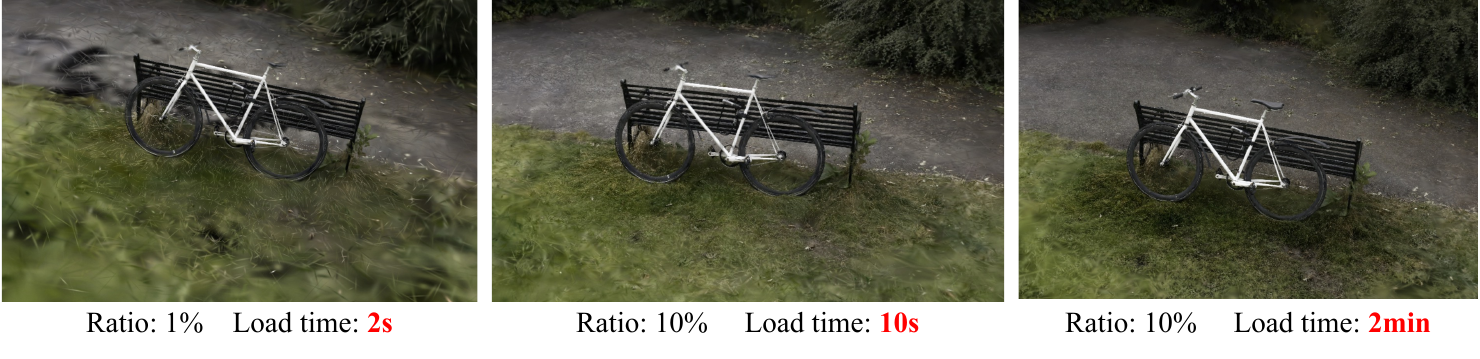}
  
\caption{Potential use of elastic inference in the application scenario of incremental scene loading.}
\label{fig:app}
\end{center}
\end{figure*}

\paragraph{How useful is the elastic inference in real application scenarios?}
In practical applications, the loading and deployment of pre-trained Gaussian models inherently demand a considerable amount of time. Moreover, as shown in Fig~\ref{fig:app}, the aggregate loading time escalates proportionally with the number of Gaussian models being deployed. Elastic inference enables the rapid deployment of lower-precision, coarse-grained models, while simultaneously maximizing rendering quality within a given resource budget.
It can further allow for the incremental loading of higher-precision, detail-rich models. This enhances the user experience of 3D scene deployment scenarios over time, like mobile gaming and online VR shopping.

\begin{table*}[t]
  \centering
  \caption{Quantitative results of FlexGS across various elastic ratios compared with other methods on Mip-NeRF360:\{\textit{bicycle}\}~\cite{mip360} (LightGS* denotes the LightGaussian without finetuning after pruning).}
  \resizebox{\textwidth}{!}{
  \setlength{\tabcolsep}{3pt}
    \begin{tabular}{c|ccc|ccc|ccc|ccc}
    \toprule
    \multirow{2}[2]{*}{Method} & \multicolumn{3}{c|}{1\%} & \multicolumn{3}{c|}{5\%} & \multicolumn{3}{c|}{10\%} & \multicolumn{3}{c}{15\%} \\
          & PSNR↑ & SSIM↑ & LPIPS↓ & PSNR↑ & SSIM↑ & LPIPS↓ & PSNR↑ & SSIM↑ & LPIPS↓ & PSNR↑ & SSIM↑ & LPIPS↓ \\
    \midrule
    LightGS*~\cite{fan2023lightgaussian} & 14.561  & 0.3327  & 0.5782  & 16.379  & 0.4445  & 0.4498  & 18.235  & 0.5388  & 0.3716  & 20.042  & 0.6207  & 0.3163  \\
    LightGS~\cite{fan2023lightgaussian} & 21.896  & 0.4814  & 0.5335  & 23.222  & 0.5952  & 0.3989  & 24.131  & 0.6769  & 0.3209  & \textbf{24.714} & 0.7230  & 0.2714  \\
    C3DGS~\cite{niedermayr2024compressed} & 21.740  & 0.4770  & 0.5350  & 23.110  & 0.5780  & 0.4110  & 23.910  & 0.6560  & 0.3390  & 24.460  & 0.7060  & 0.2930  \\
    EAGLES~\cite{Sharath} & 21.326  & 0.4549  & 0.5618  & 22.970  & 0.5600  & 0.4400  & 23.690  & 0.6300  & 0.3900  & 23.530  & 0.6300  & 0.3600  \\
    Ours  & \textbf{22.385} & \textbf{0.5350} & \textbf{0.4806} & \textbf{23.988} & \textbf{0.6865} & \textbf{0.3330} & \textbf{24.476} & \textbf{0.7302} & \textbf{0.2769} & 24.596  & \textbf{0.7408} & \textbf{0.2576} \\
    \bottomrule
    \end{tabular}%
    }
  \label{mipnerf-bicycle}%
\end{table*}%

\begin{table*}[t]
  \centering
  \caption{Quantitative results of FlexGS across various elastic ratios compared with other methods on Mip-NeRF360:\{\textit{bonsai}\}~\cite{mip360} (LightGS* denotes the LightGaussian without finetuning after pruning).}
  \resizebox{\textwidth}{!}{
  \setlength{\tabcolsep}{3pt}
    \begin{tabular}{c|ccc|ccc|ccc|ccc}
    \toprule
    \multirow{2}[2]{*}{Method} & \multicolumn{3}{c|}{1\%} & \multicolumn{3}{c|}{5\%} & \multicolumn{3}{c|}{10\%} & \multicolumn{3}{c}{15\%} \\
          & PSNR↑ & SSIM↑ & LPIPS↓ & PSNR↑ & SSIM↑ & LPIPS↓ & PSNR↑ & SSIM↑ & LPIPS↓ & PSNR↑ & SSIM↑ & LPIPS↓ \\
    \midrule
    LightGS*~\cite{fan2023lightgaussian}* & 15.231  & 0.4839  & 0.5181  & 19.071  & 0.6481  & 0.3567  & 21.774  & 0.7607  & 0.2632  & 24.025  & 0.8326  & 0.2020  \\
    LightGS~\cite{fan2023lightgaussian} & 22.397  & 0.6844  & 0.4185  & 27.359  & 0.8399  & 0.2327  & 29.374  & 0.9025  & 0.1580  & 30.590  & 0.9289  & 0.1220  \\
    C3DGS~\cite{niedermayr2024compressed} & 21.810  & 0.6640  & 0.4420  & 25.700  & 0.8020  & 0.2660  & 28.240  & 0.8810  & 0.1780  & 29.560  & 0.9150  & 0.1380  \\
    EAGLES~\cite{Sharath} & 20.601  & 0.6263  & 0.4921  & 24.660  & 0.7700  & 0.3300  & 26.420  & 0.8300  & 0.2500  & 27.450  & 0.8600  & 0.2200  \\
    Ours  & \textbf{24.420} & \textbf{0.7415} & \textbf{0.3511} & \textbf{28.449} & \textbf{0.8775} & \textbf{0.2024} & \textbf{30.605} & \textbf{0.9296} & \textbf{0.1324} & \textbf{31.606} & \textbf{0.9452} & \textbf{0.1063} \\
    \bottomrule
    \end{tabular}%
    }
  \label{mipnef-bonsai}%
\end{table*}%

\begin{table*}[t]
  \centering
  \caption{Quantitative results of FlexGS across various elastic ratios compared with other methods on Mip-NeRF360:\{\textit{counter}\}~\cite{mip360} (LightGS* denotes the LightGaussian without finetuning after pruning).}
  \resizebox{\textwidth}{!}{
  \setlength{\tabcolsep}{3pt}
    \begin{tabular}{c|ccc|ccc|ccc|ccc}
    \toprule
    \multirow{2}[2]{*}{Method} & \multicolumn{3}{c|}{1\%} & \multicolumn{3}{c|}{5\%} & \multicolumn{3}{c|}{10\%} & \multicolumn{3}{c}{15\%} \\
          & PSNR↑ & SSIM↑ & LPIPS↓ & PSNR↑ & SSIM↑ & LPIPS↓ & PSNR↑ & SSIM↑ & LPIPS↓ & PSNR↑ & SSIM↑ & LPIPS↓ \\
    \midrule
    LightGS*~\cite{fan2023lightgaussian} & 14.866  & 0.4669  & 0.5304  & 18.713  & 0.6149  & 0.3806  & 21.300  & 0.7064  & 0.3025  & 23.176  & 0.7671  & 0.2496  \\
    LightGS~\cite{fan2023lightgaussian} & 22.474  & 0.6886  & 0.4167  & 25.882  & 0.8085  & 0.2644  & 27.481  & 0.8590  & 0.2002  & 28.348  & 0.8857  & 0.1652  \\
    C3DGS~\cite{niedermayr2024compressed} & 21.920  & 0.6640  & 0.4420  & 25.820  & 0.8040  & 0.2660  & 27.390  & 0.8640  & 0.1920  & 28.120  & 0.8890  & 0.1580  \\
    EAGLES~\cite{Sharath} & 21.750  & 0.6700  & 0.4400  & 23.330  & 0.7400  & 0.3600  & 24.640  & 0.7900  & 0.2900  & 25.420  & 0.8200  & 0.2500  \\
    Ours  & \textbf{23.367} & \textbf{0.7282} & \textbf{0.3647} & \textbf{26.151} & \textbf{0.8298} & \textbf{0.2402} & \textbf{27.577} & \textbf{0.8807} & \textbf{0.1733} & \textbf{28.264} & \textbf{0.8997} & \textbf{0.1453} \\
    \bottomrule
    \end{tabular}%
    }
  \label{mipnerf-counter}%
\end{table*}%

\begin{table*}[t]
  \centering
  \caption{Quantitative results of FlexGS across various elastic ratios compared with other methods on Mip-NeRF360:\{\textit{flowers}\}~\cite{mip360} (LightGS* denotes the LightGaussian without finetuning after pruning).}
  \resizebox{\textwidth}{!}{
  \setlength{\tabcolsep}{3pt}
    \begin{tabular}{c|ccc|ccc|ccc|ccc}
    \toprule
    \multirow{2}[2]{*}{Method} & \multicolumn{3}{c|}{1\%} & \multicolumn{3}{c|}{5\%} & \multicolumn{3}{c|}{10\%} & \multicolumn{3}{c}{15\%} \\
          & PSNR↑ & SSIM↑ & LPIPS↓ & PSNR↑ & SSIM↑ & LPIPS↓ & PSNR↑ & SSIM↑ & LPIPS↓ & PSNR↑ & SSIM↑ & LPIPS↓ \\
    \midrule
    LightGS*~\cite{fan2023lightgaussian} & 12.883  & 0.2349  & 0.6578  & 15.128  & 0.3436  & 0.5264  & 16.853  & 0.4242  & 0.4596  & 18.211  & 0.4808  & 0.4204  \\
    LightGS~\cite{fan2023lightgaussian} & 18.274  & 0.3487  & 0.6104  & 19.837  & 0.4813  & 0.4680  & 20.763  & 0.5511  & 0.4090  & 21.306  & 0.5857  & 0.3776  \\
    C3DGS~\cite{niedermayr2024compressed} & 18.100  & 0.3340  & 0.6160  & 19.570  & 0.4530  & 0.4790  & 20.400  & 0.5200  & 0.4250  & 20.920  & 0.5560  & 0.3970  \\
    EAGLES~\cite{Sharath} & 17.862  & 0.3158  & 0.6461  & 19.390  & 0.4300  & 0.5300  & 20.110  & 0.4800  & 0.4700  & 20.530  & 0.5200  & 0.4400  \\
    Ours  & \textbf{18.363} & \textbf{0.3771} & \textbf{0.5656} & \textbf{21.489} & \textbf{0.5828} & \textbf{0.3836} & \textbf{21.691} & \textbf{0.5972} & \textbf{0.3646} & \textbf{21.699} & \textbf{0.5989} & \textbf{0.3581} \\
    \bottomrule
    \end{tabular}%
    }
  \label{mipnerf-flowers}%
\end{table*}%

\begin{table*}[t]
  \centering
  \caption{Quantitative results of FlexGS across various elastic ratios compared with other methods on T\&T:\{\textit{train}\}~\cite{tant} (LightGS* denotes the LightGaussian without finetuning after pruning).}
  \resizebox{\textwidth}{!}{
  \setlength{\tabcolsep}{3pt}
    \begin{tabular}{c|ccc|ccc|ccc|ccc}
    \toprule
    \multirow{2}[2]{*}{Method} & \multicolumn{3}{c|}{1\%} & \multicolumn{3}{c|}{5\%} & \multicolumn{3}{c|}{10\%} & \multicolumn{3}{c}{15\%} \\
          & PSNR↑ & SSIM↑ & LPIPS↓ & PSNR↑ & SSIM↑ & LPIPS↓ & PSNR↑ & SSIM↑ & LPIPS↓ & PSNR↑ & SSIM↑ & LPIPS↓ \\
    \midrule
    LightGS* & 11.907 & 0.3357 & 0.5643 & 15.106 & 0.5287 & 0.3958 & 17.144 & 0.6497 & 0.2905 & 18.711 & 0.7287 & 0.2264 \\
    LightGS & 18.304 & 0.5845 & 0.4506 & 21.201 & 0.7721 & 0.2476 & 22.241 & 0.8363 & 0.1682 & \textbf{22.835} & \textbf{0.8655} & \textbf{0.1298} \\
    C3DGS & 17.657 & 0.5397 & 0.4790 & 20.526 & 0.7540 & 0.2159 & 21.595 & 0.8217 & 0.1852 & 22.274 & 0.8513 & 0.1427 \\
    EAGLES & 16.715 & 0.4825 & 0.5370 & 18.794 & 0.6452 & 0.3798 & 19.633 & 0.7145 & 0.3021 & 20.175 & 0.7545 & 0.2583 \\
    Ours  & \textbf{19.189} & \textbf{0.6607} & \textbf{0.3726}  & \textbf{21.629} & \textbf{0.8042} & \textbf{0.2135} & \textbf{22.475} & \textbf{0.8519} & \textbf{0.155} & 22.830  & 0.8645 & 0.1354  \\
    \bottomrule
    \end{tabular}%
    }
  \label{tt-train}%
\end{table*}%

\begin{table*}[t]
  \centering
  \caption{Quantitative results of FlexGS across various elastic ratios compared with other methods on T\&T:\{\textit{truck}\}~\cite{tant} (LightGS* denotes the LightGaussian without finetuning after pruning).}
  \resizebox{\textwidth}{!}{
  \setlength{\tabcolsep}{3pt}
    \begin{tabular}{c|ccc|ccc|ccc|ccc}
    \toprule
    \multirow{2}[2]{*}{Method} & \multicolumn{3}{c|}{0.0100} & \multicolumn{3}{c|}{0.0500} & \multicolumn{3}{c|}{0.1000} & \multicolumn{3}{c}{0.1500} \\
          & PSNR↑ & SSIM↑ & LPIPS↓ & PSNR↑ & SSIM↑ & LPIPS↓ & PSNR↑ & SSIM↑ & LPIPS↓ & PSNR↑ & SSIM↑ & LPIPS↓ \\
    \midrule
    LightGS* & 11.449 & 0.4011 & 0.4998 & 15.287 & 0.6407 & 0.2709 & 18.501 & 0.7652 & 0.1750 & 21.179 & 0.8381 & 0.1261 \\
    LightGS & 20.641 & 0.7227 & 0.3079 & 24.993 & 0.8952 & 0.1057 & 26.478 & 0.9284 & 0.0631 & \textbf{27.130} & \textbf{0.9395} & 0.0505 \\
    C3DGS & 20.551 & 0.7201 & 0.3087 & 24.457 & 0.8877 & 0.1135 & 26.028 & 0.9241 & 0.0681 & 26.736 & 0.9360 & 0.0532 \\
    EAGLES & 18.346 & 0.6084 & 0.4331 & 21.753 & 0.7874 & 0.2317 & 23.164 & 0.8466 & 0.1624 & 24.034 & 0.8773 & 0.1282 \\
    Ours  & \textbf{22.433} & \textbf{0.8061} & \textbf{0.2248} & \textbf{25.374} & \textbf{0.9062} & \textbf{0.0934} & \textbf{26.587} & \textbf{0.9324} & \textbf{0.0579} & 27.000  & 0.9393 & \textbf{0.0501} \\
    \bottomrule
    \end{tabular}%
    }
  \label{tt-truck}%
\end{table*}%

\begin{table*}[t]
  \centering
  \caption{Quantitative results of FlexGS across various elastic ratios compared with other methods on Zip-NeRF:\{\textit{Berlin}\}~\cite{barron2023zip} (LightGS* denotes the LightGaussian without finetuning after pruning).}
  \resizebox{\textwidth}{!}{
  \setlength{\tabcolsep}{3pt}
    \begin{tabular}{c|ccc|ccc|ccc|ccc}
    \toprule
    \multirow{2}[2]{*}{Method} & \multicolumn{3}{c|}{1\%} & \multicolumn{3}{c|}{5\%} & \multicolumn{3}{c|}{10\%} & \multicolumn{3}{c}{15\%} \\
          & PSNR↑ & SSIM↑ & LPIPS↓ & PSNR↑ & SSIM↑ & LPIPS↓ & PSNR↑ & SSIM↑ & LPIPS↓ & PSNR↑ & SSIM↑ & LPIPS↓ \\
    \midrule
    LightGS* & 15.628  & 0.6578  & 0.5101  & 20.121  & 0.7457  & 0.4177  & 22.374  & 0.7951  & 0.3629  & 23.897  & 0.8278  & 0.3287  \\
    LightGS & 20.693  & 0.7432  & 0.4554  & 24.272  & 0.8194  & 0.3672  & 25.782  & 0.8549  & 0.3204  & 26.563  & 0.8722  & 0.2958  \\
    C3DGS & 20.243  & 0.7351  & 0.4679  & 23.055  & 0.7887  & 0.4047  & 24.694  & 0.8290  & 0.3491  & 25.383  & 0.8479  & 0.3238  \\
    EAGLES & 18.984  & 0.7198  & 0.4728  & 21.543  & 0.7598  & 0.4422  & 23.045  & 0.7927  & 0.4001  & 23.987  & 0.8134  & 0.3724  \\
    Ours  & \textbf{21.560} & \textbf{0.7664} & \textbf{0.4257} & \textbf{24.936} & \textbf{0.8377} & \textbf{0.3407} & \textbf{26.242} & \textbf{0.8669} & \textbf{0.3000} & \textbf{26.770} & \textbf{0.8793} & \textbf{0.2794} \\
    \bottomrule
    \end{tabular}%
    }
  \label{zipnerf-berlin}%
\end{table*}%

\begin{table*}[t]
  \centering
  \caption{Quantitative results of FlexGS across various elastic ratios compared with other methods on Zip-NeRF:\{\textit{London}\}~\cite{barron2023zip} (LightGS* denotes the LightGaussian without finetuning after pruning).}
  \resizebox{\textwidth}{!}{
  \setlength{\tabcolsep}{3pt}
    \begin{tabular}{c|ccc|ccc|ccc|ccc}
    \toprule
    \multirow{2}[2]{*}{Method} & \multicolumn{3}{c|}{1\%} & \multicolumn{3}{c|}{5\%} & \multicolumn{3}{c|}{10\%} & \multicolumn{3}{c}{15\%} \\
          & PSNR↑ & SSIM↑ & LPIPS↓ & PSNR↑ & SSIM↑ & LPIPS↓ & PSNR↑ & SSIM↑ & LPIPS↓ & PSNR↑ & SSIM↑ & LPIPS↓ \\
    \midrule
    LightGS* & 15.829  & 0.5365  & 0.5922  & 19.372  & 0.6468  & 0.4942  & 21.456  & 0.7029  & 0.4324  & 22.798  & 0.7400  & 0.3916  \\
    LightGS & 20.481  & 0.6535  & 0.5475  & 23.474  & 0.7274  & 0.4498  & 24.681  & 0.7677  & 0.3935  & 25.358  & 0.7914  & 0.3592  \\
    C3DGS & 20.162  & 0.6426  & 0.5582  & 22.843  & 0.7028  & 0.4833  & 24.032  & 0.7428  & 0.4291  & 24.644  & 0.7674  & 0.3857  \\
    EAGLES & 18.241  & 0.6136  & 0.5655  & 20.826  & 0.6638  & 0.5370  & 22.248  & 0.6943  & 0.4956  & 23.027  & 0.7164  & 0.4643  \\
    Ours  & \textbf{21.29} & \textbf{0.6697} & \textbf{0.5234} & \textbf{24.015} & \textbf{0.7458} & \textbf{0.4200} & \textbf{25.086} & \textbf{0.7834} & \textbf{0.3652} & \textbf{25.540} & \textbf{0.8012} & \textbf{0.3383} \\
    \bottomrule
    \end{tabular}%
    }
  \label{zip-london}%
\end{table*}%

\end{document}

%% file: sec/0_abstract.tex
\begin{abstract}
3D Gaussian splatting (3DGS) has enabled various applications in 3D scene representation and novel view synthesis due to its efficient rendering capabilities. However, 3DGS demands relatively significant GPU memory, limiting its use on devices with restricted computational resources. Previous approaches have focused on pruning less important Gaussians, effectively compressing 3DGS but often requiring a fine-tuning stage and lacking adaptability for the specific memory needs of different devices. In this work, we present an elastic inference method for 3DGS. Given an input for the desired model size, our method selects and transforms a subset of Gaussians, achieving substantial rendering performance without additional fine-tuning. We introduce a tiny learnable module that controls Gaussian selection based on the input percentage, along with a transformation module that adjusts the selected Gaussians to complement the performance of the reduced model. Comprehensive experiments on ZipNeRF, MipNeRF and Tanks\&Temples scenes demonstrate the effectiveness of our approach. Code is available at \href{https://flexgs.github.io/}{https://flexgs.github.io/}.
\end{abstract}

%% file: sec/1_intro.tex
\section{Introduction}

Neural scene representations have significantly advanced the field of novel view synthesis by enabling the efficient and high-fidelity reconstruction of complex 3D scenes. Among these, 3D Gaussian Splatting (3DGS)~\cite{kerbl20233d,lin2024vastgaussian, zhou2024hugs,ham2024dragon, zhang2024gs, zhang2025securegs} has emerged as a prominent technique due to its efficient rendering capabilities and ability to handle large-scale scenes. By representing scenes as a collection of anisotropic Gaussians, 3DGS strikes a balance between rendering speed and visual quality, making it suitable for real-time applications.
\begin{figure}[ht]
\begin{center}
  \includegraphics[width=\linewidth]{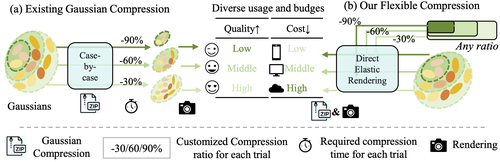}  
  \vspace{-1em}
    \caption{Comparison between compression methods for 3D Gaussian Splatting. (a) Traditional case-by-case compression requires separate fine-tuning for each ratio. (b) Our elastic inference enables dynamic compression at any ratio without re-training, adaptively balancing quality and computational costs for diverse deployment scenarios.}\vspace{-2.3em}
\label{teaser}
\end{center}
\end{figure}
Despite 3DGS being a promising technique for various applications and platforms, its widespread adoption faces two major challenges. First, standard 3DGS models are costly, requiring users to store millions of Gaussian primitives, each with attributes such as position, scaling, rotation, opacity, and color, leading to significant memory overhead. For instance, as reported in \cite{liu2025citygaussian}, deploying and rendering a city-scale scene (over 20 million Gaussians) requires at least 4 GB of networking bandwidth and GPU memory, which is not always affordable to smartphones and laptops with weak GPU. To mitigate this, recent work compresses 3DGS via pruning~\cite{fan2023lightgaussian,niedermayr2024compressed}, which employs heuristics to estimate the importance of Gaussians and subsequently prune the less important ones. The pruning process is typically followed by an additional fine-tuning stage, which requires around $16.7\%$ of the original training cost to restore rendering quality for each compression. 

Second, even efficient pruned 3DGS models often struggle to meet the diverse hardware constraints of different deployment environments. 
For instance, as a potential application of 3DGS, virtual reality city roaming and room tours require the deployment of 3DGS-represented scenes across various devices: PC-connected headsets (e.g., Valve Index\textsuperscript\textregistered) rely on consumer-grade GPUs in the connected PC; while standalone headsets (e.g., Meta Quest\textsuperscript\textregistered) feature integrated GPUs with limited memory and processing power. Additionally, different device models (such as those in the iPhone\textsuperscript\textregistered series) may have varying configurations of computing and graphics units, further increasing the diversity of computational demands.
Although one can compress an optimized 3DGS for every new device constraint, the process is inefficient as each specific computational budget must be handled separately. A more cost-effective solution is to generate a set of compressed models for popular devices and select the model that best meets the requirements of other devices. However, rendering quality may not reach the optimal level on devices for which the models are not specifically tailored, compromising the user experience.

In this work, we introduce FlexGS, which can be trained once and seamlessly adapt to varying computational constraints, eliminating the need for costly retraining or fine-tuning for each configuration / hardware constraint.
Given an input specifying the desired model size, our method selects and transforms a subset of Gaussians to meet the memory requirements while maintaining considerable rendering performance.
Through comprehensive experiments on Mip-NeRF 360 \cite{mip360}, Temples\&Tanks~\cite{tant}, and Zip-NeRF \cite{barron2023zip} datasets, we demonstrate that FlexGS can achieve competitive rendering quality with any user-desired fraction of the memory footprint of the original 3DGS models.
The key contributions of FlexGS are as follows:

\begin{itemize}
    \item FlexGS seamlessly adapts to varying memory budgets at inference time. Unlike previous methods that require separate models or extensive fine-tuning for each target size, FlexGS operates with a single model that can elastically adjust the number of active Gaussians based on real-time input, offering significant flexibility for deployment on  devices with different performance budgets.
      
    \item We propose a learnable module that dynamically selects the most significant Gaussians based on an input compression ratio. This selector is guided by a novel Global Importance (GI) metric, which quantifies the contribution of each Gaussian to the overall rendering quality by considering factors such as spatial coverage, opacity, and transmittance.
    
    \item To compensate for the removal of Gaussians and mitigate potential performance degradation, we introduce a transformation module that adjusts the attributes of the selected Gaussians. This module predicts spatial and geometric transformations, ensuring that the reduced set of Gaussians can effectively represent the scene across different ratios without overfitting to specific configurations.
    
\end{itemize}

%% file: sec/2_related.tex
\section{Related Work}
\paragraph{Compact Neural Radiance Field.} Neural Radiance Field (NeRF)~\cite{NeRF} revolutionized novel view synthesis by introducing an implicit MLP model that generates novel views via ray-based RGB accumulation. While groundbreaking, its computational demands from dense sampling and complex networks limit real-time applications. Various solutions like Instant-NGP~\cite{INGP}, TensoRF~\cite{TensoRF}, K-planes~\cite{K-planes}, and DVGO~\cite{DVGO} proposed explicit grid representations to accelerate computation, though at increased storage costs.
To address storage efficiency, two main compression paradigms have emerged. Direct value compression includes techniques like pruning~\cite{VQRF, Re:NeRF}, vector quantization~\cite{VQRF, CompactNeRF}, and entropy-based methods in BiRF~\cite{BiRF} and SHACIRA~\cite{SHACIRA}. Structure-aware approaches leverage spatial correlations through wavelet transforms~\cite{MaskDWT}, rank decomposition~\cite{CCNeRF}, or predictive coding~\cite{SPC-NeRF}, exploiting the inherent regularity of feature grids. CNC~\cite{cnc2024} demonstrates the effectiveness of structural compression through significant rate-distortion improvements.

\paragraph{Compact 3D Gaussian Splatting.} 
3D Gaussian Splatting (3DGS)~\cite{3DGS} presents an alternative to NeRF by modeling scenes with optimizable 3D Gaussians. Through differentiable splatting and efficient rasterization~\cite{raster}, it achieves faster training and rendering while maintaining visual quality. However, the method typically requires millions of Gaussian primitives, each storing independent spatial and appearance attributes, resulting in substantial memory requirements. The scattered nature of point-based representation poses unique challenges for structural compression. Recent advances in compressing 3DGS can be categorized into three main approaches scene ~\cite{bagdasarian20243dgs}{:} compaction~\cite{lee2024compact, fan2023lightgaussian, kim2024color}, attribute compression~\cite{lee2024compact, niedermayr2024compressed, fan2023lightgaussian, navaneet2023compact3d}, and structured representation~\cite{lu2024scaffold, morgenstern2023compact}.  

\textbf{Scene compaction} in 3DGS primarily follows two approaches: densification and pruning. Densification methods strategically add Gaussians to improve scene representation. For example, the Color-cued Efficient Densification method~\cite{kim2024color} leverages view-independent spherical harmonics coefficients to enhance detail capture while keeping minimum densification. In contrast, pruning provides a more direct path to compactness by removing redundant Gaussians. Compact3DGS~\cite{lee2024compact} and RDO-Gaussian~\cite{wang2024end} introduce mask promoters for iterative Gaussian elimination, while LightGaussian~\cite{fan2023lightgaussian} employs a one-time importance scoring system for efficient pruning. 

\textbf{Attribute compression} reduces per-Gaussian storage requirements through quantization. Several methods~\cite{navaneet2024compgs, niedermayr2024compressed,fan2023lightgaussian} have introduced vector quantization to compress Gaussian parameters. Reduced3DGS~\cite{papantonakis2024reducing} takes a different approach by proposing adaptive adjustment of spherical harmonics degree based on view-dependent effects. However, these compression techniques often introduce additional computational overhead during rendering. 

\textbf{Structured representation} methods organize Gaussians to exploit spatial coherence. Scaffold-GS~\cite{lu2024scaffold} presents an anchor-based representation where attributes are associated with representative points rather than individual Gaussians. Gaussian Grids~\cite{morgenstern2023compact} explores novel approaches through 2D spatial organization, though spatial redundancy remains partially unexploited in current methods. While existing methods rely on fixed compression ratios or model-specific optimization, our work introduces a unified framework that dynamically adapts to varying computational constraints. By jointly optimizing scene compactness and spatial relationships, our approach enables efficient deployment across diverse hardware platforms through a single trained model.

\paragraph{Elastic Inference.}
Flexible inference has been a major focus of research, particularly in convolutional neural networks (CNNs). \citet{yu2018slimmable, yu2019universally} pioneered slimmable neural networks, allowing a single model to operate with different numbers of convolutional kernels. OFA~\citep{cai2019once} further advanced this idea by generalizing pruning techniques to build a model capable of adapting to multiple configurations. 

More recently, slimmable models have been extended to Transformer architectures. \citet{kusupati2022matryoshka} proposed the nested Matryoshka structure, while Matformer~\citep{kudugunta2023matformer} applied this concept to the MLP hidden dimensions of large language models. Flextron~\citep{cai2024flextron} introduced elastic multi-head attention and constructed elastic LLMs using learnable routers. However, these methods are designed specifically for neural networks and cannot be directly applied to 3DGS.

%% file: sec/3_method.tex
\section{Method}

\begin{figure*}[ht]
\label{fig:method}
\begin{center}
  \includegraphics[width=0.9\textwidth]{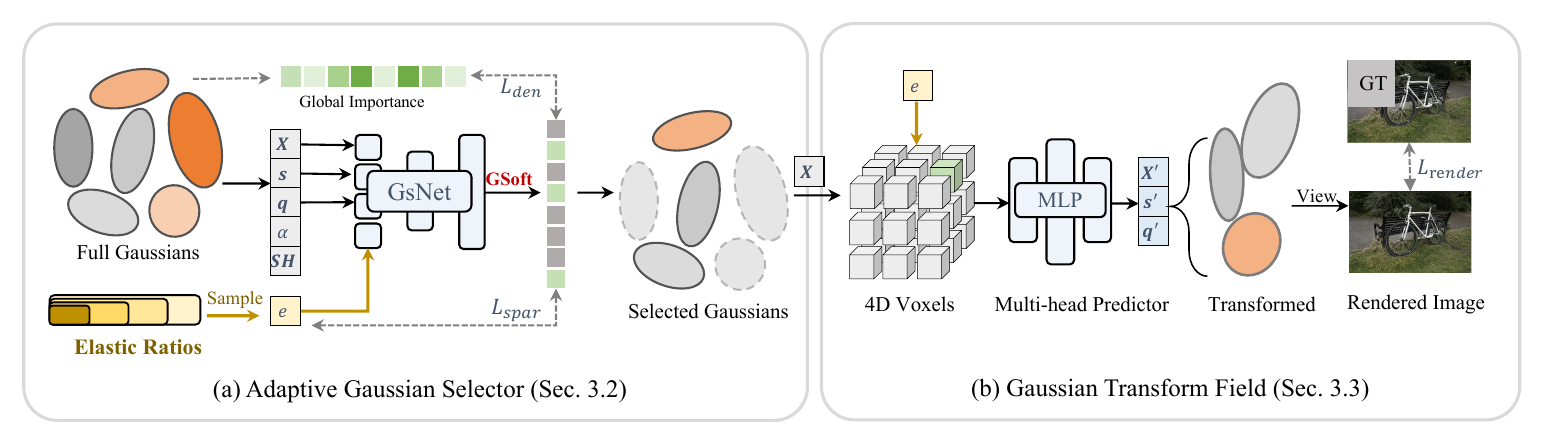}
\vspace{-1em}
\caption{
\textbf{Overall framework of FlexGS}: (a) Adaptive Gaussian Selector: utilizes a GsNet to output the differentiable mask and Global Important as guidance for adaptive selection; (b) Gaussian Transform Field: queries a Spatial-Ratio Neural Field and outputs the transforms of Gaussian attributes under the given elastic ratio. All the modules are jointly optimized to minimize the rendering loss. 
}\vspace{-2em}
\label{fig:method}
\end{center}
\end{figure*}

Given a set of ratios $\boldsymbol{E}=\{e_i\}_{i=1}^{n}$, our proposed elastic framework, FlexGS, trains 3DGS once and enables it to render with an arbitrary number of Gaussians without requiring additional fine-tuning.
FlexGS consists of two modules: 1) A learnable Gaussian selector for adaptively selecting Gaussians under a given ratio (Sec.~\ref{sec3-2}); 2) A ratio-dependent Gaussian transformation field for displacing selected Gaussians to compensate missing details due to the decrease of the number of Gaussians (Sec.~\ref{sec3-3}). In the following Sec.~\ref{sec3-1}, we first overview the basics of 3DGS.

\subsection{Preliminary}\label{sec3-1}
3DGS~\cite{kerbl20233d} utilizes a set of anisotropic Gaussians with various attributes for 3D representation. 
As described in Eq.~\ref{gaussian define}, each Gaussian is defined by a mean value $\boldsymbol{X}$ and a covariance matrix $\boldsymbol{\Sigma}$, which collectively delineate its spatial position and size. For differentiable optimization, $\boldsymbol{\Sigma}$ can be decomposed into a scaling matrix $\mathbf{S}$ 
and a rotation matrix $\mathbf{R}$. A 3D vector $\boldsymbol{s}$ for scaling and a quaternion $\boldsymbol{q}$ for rotation are utilized for independent optimization of both factors. 
\begin{equation}
\label{gaussian define}
\mathrm{G}(\boldsymbol{X})=e^{-\frac{1}{2}\boldsymbol{X}^\top\boldsymbol{\Sigma}^{-1}\boldsymbol{X}}, \boldsymbol{\Sigma} = \mathbf{R}\mathbf{S}\mathbf{S}^\top\mathbf{R}^\top.
\end{equation}

When rendering 2D images,  differential splatting~\cite{yifan2019differentiable} is employed for the 3D Gaussians within the camera planes. The covariance matrix $\boldsymbol{\Sigma'}$ in camera coordinates can be calculated as $\boldsymbol{\Sigma}^{\prime} = \boldsymbol{JW}\boldsymbol{\Sigma} \boldsymbol{W}^\top\boldsymbol{J}^\top$, where $\boldsymbol{W}$ is the viewing transform matrix and $\boldsymbol{J}$ is the Jacobian matrix of the affine approximation of the projective transformation. For each pixel, the color $\boldsymbol{c}$ and opacity $\alpha$ of Gaussians are used for the blending of $N$ ordered points that overlap the pixel:
\begin{equation}
\label{color-blending}
\boldsymbol{C} = \sum_{i\in N}\boldsymbol{c_i} \alpha_i \prod_{j=1}^{i-1} (1-\alpha_j).
\end{equation}

In summary, each Gaussian is characterized by: position $\boldsymbol{X} \in \mathbb{R}^3$, scaling vector $\boldsymbol{s} \in \mathbb{R}^3$, rotation vector $\boldsymbol{q} \in \mathbb{R}^3$, opacity $\alpha \in \mathbb{R}$ and color $\boldsymbol{c}$ defined by spherical harmonics coefficients $\text{SH} \in \mathbb{R}^{{(d+1)}^2}$ ($d$ is the $\text{SH}$ degree).

\subsection{Adaptive Gaussian Selector}
\label{sec3-2}

\paragraph{Background and Problem Formulation.} 
For a 3DGS with $N$ Gaussians and a target ratio $e$ (representing the desired percentage of remaining Gaussians), FlexGS aims to predict a binary scaler $\lambda_i$ for each Gaussian.
This scaler determines whether the $i$-th Gaussian is selected ($\lambda_i=1$) or not ($\lambda_i=0$), ensuring minimal performance degradation while satisfying the constraint $\sum_i \lambda_i =eN$.

To solve this problem, recent post-hoc compression methods~\cite{fan2023lightgaussian, niedermayr2024compressed} use heuristics to assign importance scores $s_i$ based on the attributes of each Gaussian $i$. They retain only those Gaussians with the highest scores, i.e., $\lambda_i = 1 $ if $ i $ is among the Top-$K$ Gaussians ranked by $ s_i $, where $K = eN $.
However, these methods require multiple separate runs for each ratio $e$ after the Gaussian model is pre-trained. In contrast, we aim to ``elastic-ize'' 3DGS during the training, thus the Gaussian model can be inferenced with arbitrary remaining raio.
To achieve this, importance scores must be frequently re-computed as Gaussian attributes constantly evolve during training.
In addition, a fixed heuristic selection rule may fail to generalize in extreme cases, e.g., when selecting $1\%$ of Gaussians. To address this, we jointly train a learnable module to adaptively select points based on the Gaussian attributes and the desired ratio.

\paragraph{Adaptive Selection via Gumbel-Softmax.}
Due to the discrete nature of variable $\lambda_i, i\in[N]$, traditional gradient descent methods are ineffective, as they rely on the continuous differentiability of the objective function with respect to its variables. Direct optimization over discrete choices would require exhaustive search or evolutionary algorithms, both of which are computationally prohibitive due to the large search space.
To overcome this challenge, we employ Gumbel-Softmax~\cite{maddison2016concrete,jang2016categorical}, which provides a continuous relaxation of the discrete optimization problem. This technique approximates categorical distributions over discrete variables by a differentiable softmax function with Gumbel noise, enabling the use of gradient-based optimization. 

We first represent each binary variable $\lambda_i$ as a Bernoulli distribution, and instead of directly selecting between 0 and 1, we model the probability of each outcome.
Specifically, given the $i$-th Gaussian's attributes $\boldsymbol{A}_i=\{\boldsymbol{X},\boldsymbol{s},\boldsymbol{q}\}$ and a ratio $e$, we model the categorical distribution with GsNet (detailed later) as below:

\begin{equation}
\begin{aligned}
    &P(\lambda_i = m \vert \boldsymbol{A}_i, e) = \text{GsNet}(\boldsymbol{A}_i, e), \quad m \in \{0, 1\},\\
    & \text{s.t.} \sum_{m \in \{0,1\}}P(\lambda_i = m \vert \boldsymbol{A}_i, e) =1.
\end{aligned}
\end{equation}
We perform the stochastic sampling based on the categorical distribution $P(\lambda_i = m \vert \boldsymbol{A}_i, e)$ during training. 
Leveraging the differentiability of the Gumbel-Softmax, the algorithm dynamically adjusts the probabilities, assigning higher values to $m$ that are more likely to yield improved performance, reflecting the favorable selecting decisions.

\paragraph{GsNet Structure and Optimization.} 
We convert the ratio $e$, to feature $\boldsymbol{h}_e$ with a lightweight network as below:
\begin{equation}
    \boldsymbol{h}_e = \sigma(\sigma(e \boldsymbol{w}^1_e) \boldsymbol{W}^2_e),
\end{equation}
where the $\boldsymbol{w}^1_e \in \mathbb{R}^{D}$, and $\boldsymbol{W}^2_e \in \mathbb{R}^{D\times D}$. $\sigma(\cdot)$ denotes the non-linear activation function, and we adopt ReLU activation during implementation. Similarly, to embed Gaussian attributes $\boldsymbol{A}_i$, we first concatenate the elements in $\boldsymbol{A}_i$ to $\boldsymbol{\xi} \in\mathbb{R}^{H}$, and obtain $\boldsymbol{h}^a_i$ as follows: 
\begin{equation}
    \begin{aligned}
        \boldsymbol{h}^a_i &= \sigma\left(\sigma(\boldsymbol{\xi}\boldsymbol{W}^1_a)\boldsymbol{W}^2_a\right),\\
        \boldsymbol{\xi} &= \text{Concat}(\{\boldsymbol{a}: \boldsymbol{a}\in \boldsymbol{A}_i\}),
    \end{aligned}
\end{equation}
where the $\boldsymbol{W}^1_a \in \mathbb{R}^{H\times D}$, and $\boldsymbol{W}^2_a \in \mathbb{R}^{D\times D}$. 
After obtaining $\boldsymbol{h}_e$ and $\boldsymbol{h}^a_i$, features are first concatenated then mixed through a mixing layer with weight $\boldsymbol{W}\in \mathbb{R}^{2D\times 2}$, which projects the concatenated features of dimension $2D$ into a 2-dimensional output, producing logits $\boldsymbol{z}$. For a ratio $e$: 
\begin{equation}
        \boldsymbol{z} = [\boldsymbol{h}_e, \boldsymbol{h}^a_i]\boldsymbol{W}.
\end{equation}
The logits are then processed by a Gumbel-Softmax~\cite{jang2016categorical} layer to model the categorical distribution.
\begin{equation}
P(\lambda_i = m \vert \boldsymbol{A}_i, e) = \frac{\exp\left(\left(z_m + g_m\right)/\tau\right)}{\sum_{m=0,1} \exp\left(\left(z_m + g_m\right)/{\tau}\right)},
\end{equation}
where $g_m$ is sampled from $\mathrm{Gumbel}(0,1)$ distribution, and $\tau$ is a temperature parameter that controls the smoothness of the approximation. 
During the 3DGS optimization stage, binary masks $\widehat{\boldsymbol{M}}$ are sampled from the categorical distribution to select the active Gaussians for training. The temperature parameter \(\tau\) is exponentially decayed, and as \(\tau\) approaches 0, the distribution converges to a one-hot vector, allowing GsNet to make more deterministic selections.

\paragraph{Global Importance.}
Our pilot study reveals that GsNet struggles to converge smoothly due to insufficient constraints in Gaussian selection at high ratios, where there are numerous ways to mask Gaussians. Thus, introducing a mechanism to guide and prioritize Gaussian selection is necessary.
Inspired by \cite{fan2023lightgaussian, niedermayr2024compressed} that employ opacity and scale vectors to quantify the significance of Gaussian components, we introduce the Global Importance ($\text{GI}$) score as a preference criterion for selecting Gaussians.
In Eq.~\ref{GI-cal}, the Global Importance (GI) score of each Gaussian is quantified by its volumetric extent, the frequency of its projection intersecting image pixels, its opacity, and the transmittance of incident rays it attenuates:
\begin{equation}
\label{GI-cal}
\text{GI}_i = \sum_{p=1}^{QHW}{ \mathbbm{1}(G(\boldsymbol{X}_i), \boldsymbol{r}_{p}) \cdot \gamma(\boldsymbol{s}_i) \cdot \alpha_i \cdot {T}(i, p)},
\end{equation}
where $Q$, $H$, and $W$ denote the number of seen views, image height, and width. $\mathbbm{1}(G(\boldsymbol{X}_i), \boldsymbol{r}_{p})$ indicates whether the projection of $G(\boldsymbol{X}_i)$ on the camera plane has overlaps with pixel $p$.
Following~\cite{fan2023lightgaussian}, $\gamma(\boldsymbol{s}_i)$ calculates the volume normalized by the $90\%$ of the largest volume.
$T(i, p)$ is the transmittance of the $i$-th Gaussian for rendering pixel $p$. Let $\mathrm{ID}(i, p)$ denote the index of the $i$-th Gaussian among the sorted set of $N_p$ Gaussians overlapping with $p$. Then, $\gamma(\boldsymbol{s}_i)$ and $T(i, p)$ can be expressed as:
\begin{equation}
\label{cal-vol}
\gamma(\boldsymbol{s}_i)=\max(V(\boldsymbol{s}_i)/V_{90\%}, 1), V(\boldsymbol{s})=|s_1s_2s_3|\pi/3.
\end{equation}
\begin{equation}
\label{cal-t}
{T}(i, p)=\prod_{t=1}^{\mathrm{ID}(i, p)-1} (1-\alpha_t).
\end{equation}
With GI computed for each Gaussian, we select the top $\lfloor eN \rfloor$ Gaussians with the highest GI scores to obtain the guidance mask $\boldsymbol{M}_{GI}$. To incorporate this guidance and ensure the number of selected Gaussians aligns with the desired elastic ratio $e$, we regularize the selection process by closing the gap between $\widehat{\boldsymbol{M}}$ and the guidance mask via $\mathcal{L}_{GI}$, and imposing a sparsity ratio constraint via $\mathcal{L}_{spar}$:
\begin{equation}
\label{eq:supervise_select}
\mathcal{L}_{GI}=\left|\widehat{\boldsymbol{M}} -\boldsymbol{M}_{GI}\right|, 
\mathcal{L}_{spar}=\left|e - \sum_{m\in \hat{\boldsymbol{M}}}\frac{m}{N}\right|.
\end{equation}

\begin{figure}[t]
\begin{center}
  \includegraphics[trim={10 12 0 1},clip,width=0.47\textwidth]{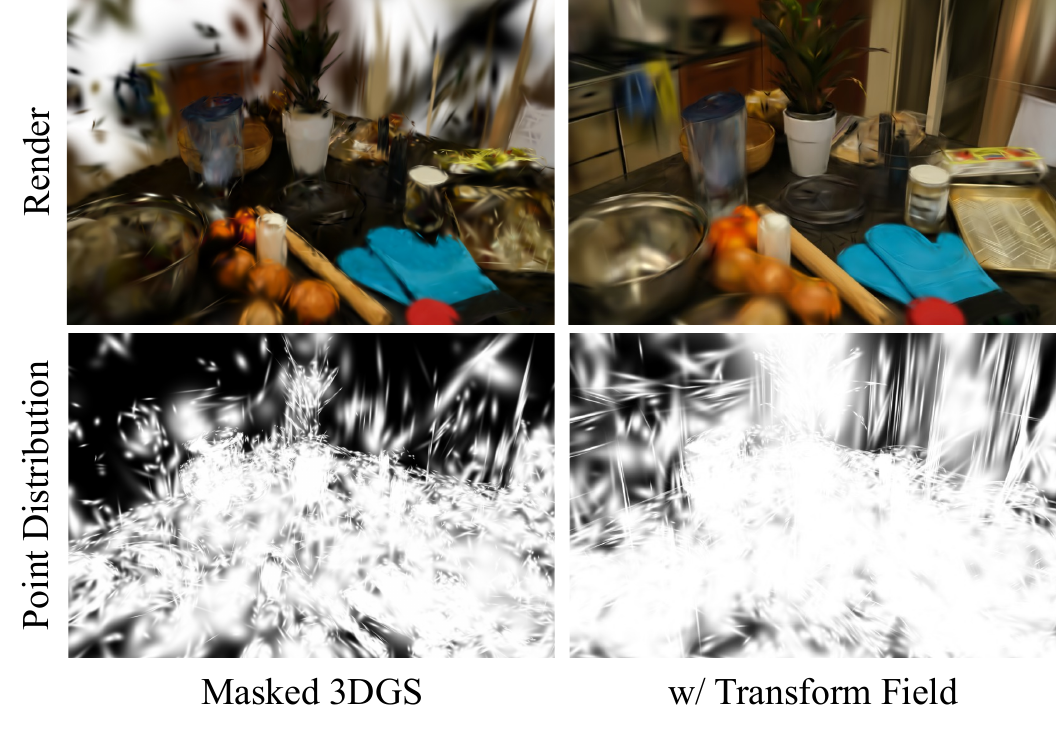}
\caption{Visualization of the point distribution for masked 3DGS (left) and FlexGS (right). Simply masking out Gaussians leads to missing areas. 
The point distribution on the left highlights the absence of points in the background.
}
\vspace{-2em}
\label{pic:viewer}
\end{center}
\end{figure}

\subsection{Gaussian Transform Field}
\label{sec3-3}
Merely optimizing Gaussian selection for an elastic ratio may lead to excessive updates of crucial Gaussians at lower ratios, thereby inducing overfitting to these ratios and subsequently diminishing rendering quality under other ratios. Furthermore, selecting a portion of 3DGS will inevitably lead to holes and missing details in the rendering results.  As visualized in Fig. \ref{pic:viewer}, simply fine-tuning the SH and opacity attributes is unlikely to compensate for those degraded areas in the masked 3DGS. 
Therefore, we propose to transform the positions, scaling, and rotations of selected Gaussians to patch the missing areas. To do this, we introduce the Gaussian Transform Field, consisting of a Spatial-Ratio Neural Field $\boldsymbol{\psi}$ and a Multi-Head Predictor $\boldsymbol{\phi}$. This module predicts position and shape transformations for each Gaussian, mapping them from their original state, where all Gaussians are selected, to the adapted state under a given elastic ratio.

\paragraph{Spatial-Ratio Neural Field.}
Inspired by previous work in dynamic scene representation \cite{wu20244d,cao2023hexplane,fridovich2023k}, our Spatial-Ratio Neural Field $\boldsymbol{\psi}$ is a mapping from 4D coordinate $(\boldsymbol{X}, e)$ to multi-channel features. Given a Gaussian at $\boldsymbol{X}$ and an elastic ratio $e$, the mapping can be queried to retrieve a ratio-dependent spatial feature for the Gaussian.
The Spatial-Ratio Neural Field in our method is modeled as a multi-resolution 4D volume: $\boldsymbol{\psi} = \{\boldsymbol{\psi}_l \in \mathbbm{R}^{4}\}_{l\in\{1,2\}}$, where each vertex in the volume stores a multi-channel feature. 
Following~\cite{wu20244d}, we factorize $\boldsymbol{\psi}_l$ into 6 planes $\boldsymbol{\psi}_l(k_1, k_2) \in \mathbb{R}^{d_{k_1} \times d_{k_2}}, (k_1,k_2) \in\{(x,y), (x,z), (x,t), (y,z), (y,t), (z,t)\}$ for efficiency consideration. In specific, the feature querying can be formulated as Eq.~\ref{eq:query}.
\begin{equation}
\label{eq:query}
\boldsymbol{f} = \mathcal{F}_m(\boldsymbol{f}_m), \quad \boldsymbol{f}_m = \bigcup_{l} \mathrm{BiInterp}\left(\boldsymbol{\psi}_l(\boldsymbol{X}, e)\right),
\end{equation}
where $\mathrm{BiInterp}$ denotes bilinear interpolation employed for interpolating vertex features stored in the multi-resolution 4D volume, and $\mathcal{F}_m$ is an MLP that fuses features across multi-resolutions.

\paragraph{Multi-Head Predictor.}
A set of multi-layer perceptrons (MLPs), denoted as $\{\boldsymbol{\phi}_X, \boldsymbol{\phi}_s, \boldsymbol{\phi}_q\}$, then transforms the queried features into displacement vectors of Gaussian attributes between their ``all-selected'' states (i.e., $e=1$) and the adapted states under the specified elastic ratio.
\begin{equation}
\label{eq:pred}
\boldsymbol{T}_X = \boldsymbol{\phi}_X(\boldsymbol{f}), \boldsymbol{T}_s = \boldsymbol{\phi}_s(\boldsymbol{f}), \boldsymbol{T}_q = \boldsymbol{\phi}_q(\boldsymbol{f}),
\end{equation}
The selected Gaussians are then transformed by the predicted displacement vectors to new positions and shapes: $
\boldsymbol{X} = \boldsymbol{X} + \boldsymbol{T}_X, 
\boldsymbol{s} = \boldsymbol{s} + \boldsymbol{T}_s, 
\boldsymbol{q} = \boldsymbol{q} + \boldsymbol{T}_q
$. 
Ultimately, we acquire a set of Gaussians that conform to the specified elastic ratios in quantity, with their spatial positions and shape adjusted accordingly to guarantee rendering quality.

\begin{table*}[htbp]
  \centering
  \caption{Quantitative results of FlexGS across various elastic ratios compared with other methods on Mip-NeRF360 dataset~\cite{mip360} (LightGS* denotes the LightGaussian without fine-tuning after pruning).}
  \resizebox{\textwidth}{!}{%
    \begin{tabular}{c|ccc|ccc|ccc|ccc}
    \toprule
    \multirow{2}{*}{Method} & \multicolumn{3}{c|}{1\%} & \multicolumn{3}{c|}{5\%} & \multicolumn{3}{c|}{10\%} & \multicolumn{3}{c}{15\%} \\
    
          & PSNR  & SSIM  & LPIPS & PSNR  & SSIM  & LPIPS & PSNR  & SSIM  & LPIPS & PSNR  & SSIM  & LPIPS \\
    \midrule
    LightGS*~\cite{fan2023lightgaussian} & 14.455  & 0.3741  & 0.5780 & 17.302  & 0.5055  & 0.4369 & 19.595  & 0.6085  & 0.3506 & 21.467  & 0.6805  & 0.2940 \\
    LightGS~\cite{fan2023lightgaussian}  & 22.265  & 0.6128  & 0.4384 & 25.223  & 0.8381  & 0.3305 & 26.570  & 0.7831  & 0.2399 & \textbf{27.239} & \textbf{0.8079} & \textbf{0.2070} \\
    C3DGS~\cite{niedermayr2024compressed}, & 21.766  & 0.5440  & 0.5101 & 24.224  & 0.7333  & 0.3452 & 25.609  & 0.7700  & 0.2457 & 26.447  & 0.7807  & 0.2277 \\
    EAGLES~\cite{Sharath} & 21.761  & 0.5173  & 0.5479 & 23.421  & 0.7389  & 0.3544 & 24.489  & 0.6822  & 0.3111 & 25.078  & 0.6789  & 0.2744 \\
    Ours  & \textbf{22.730} & \textbf{0.6128} & \textbf{0.4384} & \textbf{25.412} & \textbf{0.7475} & \textbf{0.2935} & \textbf{26.577} & \textbf{0.7933} & \textbf{0.2320} & 27.035  & 0.8072  & 0.2087 \\
    \bottomrule
    \end{tabular}%
  }
  \label{tab:sota}%
  \vspace{-0.75em}
\end{table*}%

\begin{figure*}[ht]
\begin{center}
  \includegraphics[trim={7 0 7 0},clip,width=0.92\textwidth]{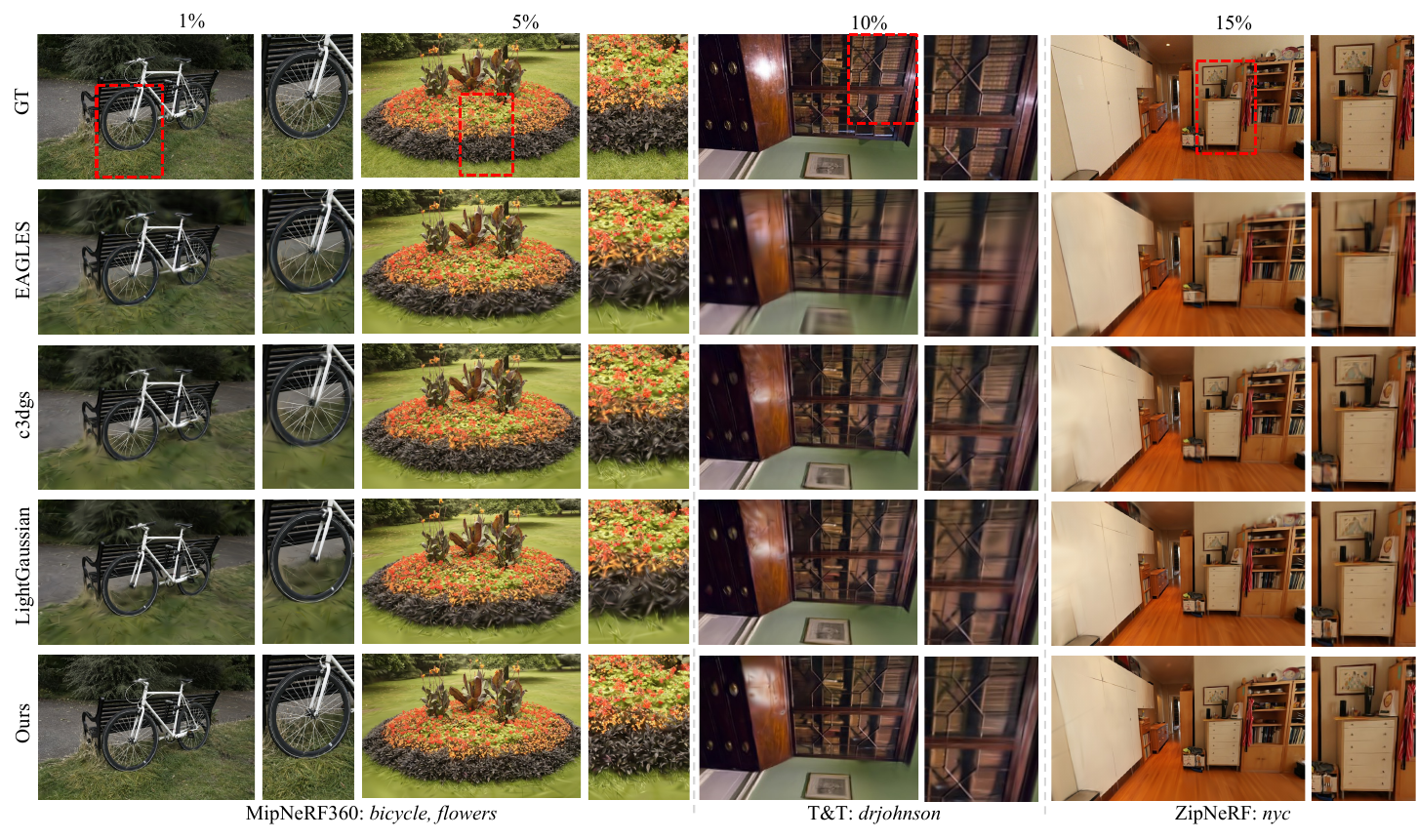}
  \vspace{-1em}
\caption{
Visual results compared with other methods on various elastic ratios:\{0.1,0.5,1.0,1.5\} on Mip-NeRF360~\cite{mip360}:~\{\textit{bicycle}, \textit{flowers}\}, T\&T~\cite{tant}:~\{\textit{drjohnson}\} and ZipNeRF~\cite{barron2023zip}:~\{\textit{nyc}\}.
}\vspace{-2em}
\label{fig:main}
\end{center}
\end{figure*}

\subsection{Optimization}
During training, to ensure differentiable Gaussian selection,
the opacity of all Gaussian is updated as follows:
\begin{equation}
\label{eq:opacity}
\alpha_i = \alpha_i \cdot \hat{\boldsymbol{M}}_i,
\end{equation}
where unselected Gaussians are rendered in transparent. At each training iteration, an elastic ratio $e$ is randomly sampled from $\boldsymbol{E}$. Subsequently, based on the output masks of our Adaptive Gaussian Selector, the top $e$ most significant Gaussians are selected from the entire Gaussians. 
At each optimization step, two images are rendered:  $\boldsymbol{I}_s^{e}$ using selected Gaussians, and $\boldsymbol{I}_f$ using all Gaussians. Both will be supervised against the ground truth $\boldsymbol{I}_{GT}$:
\begin{equation}
\label{eq:render-loss}
\mathcal{L}_{render}=|\boldsymbol{I}_s^{e}-\boldsymbol{I}_{GT}| + |\boldsymbol{I}_f - \boldsymbol{I}_{GT}|, \quad e \sim \boldsymbol{E}.
\end{equation}

The final loss function is given in Eq.~\ref{eq:final}, where $\beta_{1}$ and $\beta_{2}$ are weighting coefficients for the regularization with GI guidance and the ratio constraint, respectively.
\begin{equation}
\label{eq:final}
\mathcal{L}=\mathcal{L}_{render} + \beta_{1}\mathcal{L}_{GI} + \beta_{2}\mathcal{L}_{spar}.
\end{equation}

%% file: sec/4_exp.tex
\begin{table*}[htbp]
  \centering
  \caption{Performance of FlexGS across various elastic ratios compared with other methods on Real-world 3D datasets: T\&T~\cite{tant} and ZipNeRF~\cite{barron2023zip}. LightGS* denotes the LightGaussian without fine-tuning after pruning.} 
  \vspace{-0.5em}

\resizebox{\textwidth}{!}{
  \setlength{\tabcolsep}{3pt}
    \begin{tabular}{c|cccccccccccccccc}
    \toprule
    \multirow{3}[4]{*}{Method} & \multicolumn{8}{c}{T\&T}                                     & \multicolumn{8}{c}{Zip-NeRF} \\
\cmidrule(lr){2-9} \cmidrule(lr){10-17}       & \multicolumn{2}{c}{1\%} & \multicolumn{2}{c}{5\%} & \multicolumn{2}{c}{10\%} & \multicolumn{2}{c}{15\%} & \multicolumn{2}{c}{1\%} & \multicolumn{2}{c}{5\%} & \multicolumn{2}{c}{10\%} & \multicolumn{2}{c}{15\%} \\
\cmidrule(lr){2-3}\cmidrule(lr){4-5}\cmidrule(lr){6-7}\cmidrule(lr){8-9}
\cmidrule(lr){10-11}\cmidrule(lr){12-13}\cmidrule(lr){14-15}\cmidrule(lr){16-17}
          & PSNR↑ & SSIM↑ & PSNR↑ & SSIM↑ & PSNR↑ & SSIM↑ & PSNR↑ & SSIM↑ & PSNR↑ & SSIM↑ & PSNR↑ & SSIM↑ & PSNR↑ & SSIM↑ & PSNR↑ & SSIM↑ \\
    \midrule
    LightGS*~\cite{fan2023lightgaussian} & 14.653  & 0.4674  & 19.545  & 0.6705  & 21.724  & 0.7705  & 23.550  & 0.8271  & 14.744  & 0.5442  & 18.772  & 0.6514  & 20.913  & 0.7119  & 22.839  & 0.7506  \\
    LightGS~\cite{fan2023lightgaussian} & 22.795  & 0.7231  & 25.774  & 0.8605  & 26.820  & 0.8954  & 27.782  & 0.9098  & 19.812  & 0.6527  & 22.899  & 0.7356  & 24.223  & 0.7788  & 24.957  & 0.8040  \\
    C3DGS~\cite{niedermayr2024compressed} & 21.535  & 0.6930  & 25.625  & 0.8460  & 26.823  & 0.8853  & 26.860  & 0.9013  & 19.955  & 0.6405  & 22.105  & 0.7065  & 23.425  & 0.7518  & 24.053  & 0.7760  \\
    EAGLES~\cite{Sharath} & 19.558  & 0.6140  & 22.738  & 0.7620  & 24.058  & 0.8155  & 24.808  & 0.8440  & 18.495  & 0.6193  & 20.599  & 0.6721  & 21.934  & 0.7086  & 22.785  & 0.7359  \\
    Ours  & \textbf{23.090} & \textbf{0.7731} & \textbf{26.689} & \textbf{0.8764} & \textbf{27.069} & \textbf{0.9045} & \textbf{27.802} & \textbf{0.9101} & \textbf{20.963} & \textbf{0.6722} & \textbf{23.513} & \textbf{0.7572} & \textbf{24.691} & \textbf{0.7962} & \textbf{25.167} & \textbf{0.8132} \\
    \bottomrule
    \end{tabular}%
    }
  \label{tab:sota2}
  \vspace{-0.5em}
\end{table*}%

\section{Experiments}
\subsection{Evaluation Setup}
\paragraph{Datasets.} Our experiments focus on three widely used datasets for real-world, large-scale scene rendering: Mip-NeRF360~\cite{mip360}, Zip-NeRF~\cite{barron2023zip}, and Tanks and Temples~\cite{tant} (T\&T). Mip-NeRF360~\cite{mip360} comprises nine distinct scenes that encompass both expansive outdoor scenes and intricate indoor settings. 
Zip-NeRF~\cite{barron2023zip} consists of four sizable indoor scenes, each containing 990-1800 images. 
In our evaluation of Zip-NeRF, we adopt the identical split strategy employed for Mip-NeRF360. 
We evaluate selected scenes from the T\&T~\cite{knapitsch2017tanks}, following the same choices in~\cite{kerbl20233d}.

\paragraph{Baselines and Metrics.}
We evaluate the proposed FlexGS against state-of-the-art Compression GS approaches LightGaussian~\cite{fan2023lightgaussian} (and its variant without post-pruning fine-tuning, dubbed LightGS*), C3DGS~\cite{niedermayr2024compressed}, and EAGLES~\cite{Sharath} across various elastic ratios.
The rendering quality is meticulously evaluated using standard image quality metrics: PSNR to measure pixel-wise reconstruction fidelity, SSIM to assess structural similarity, and LPIPS~\cite{zhang2018unreasonable} to evaluate perceptual quality.

\paragraph{Implementation Details.}
The training process of FlexGS contains two stages. The initial stage comprises $15$k iterations, replicating the original 3DGS training procedure. In the subsequent stage, both GsNet and the Transform Field are optimized over $20$k iterations, with the elastic ratio randomly sampled from \{$0.01$, $0.05$, $0.10$, $0.15$\} at each iteration. In this phase, $\beta_s$ is set to $1.0$ and $\beta_d$ is configured to $0.01$. To ensure the accuracy of guidance, the Global Importance of all Gaussians is updated every $1$k iterations.
All experiments are conducted on an NVIDIA A100 GPU.

\begin{table*}[htbp]
  \centering
  \caption{Ablation studies of FlexGS on the T\&T~\cite{tant} dataset: G.I. denotes the guidance with Global Importance, Adap. Sel. denotes the Adaptive Selection, and Heuristic Sel. directly conducts elastic inference without Adaptive Selection and Transform field.} 
  \vspace{-0.5em}
  \label{tab:ablation}
  \setlength{\tabcolsep}{2.9pt} %
\resizebox{\textwidth}{!}{
    \begin{tabular}{l|cccccccccccc}
    \toprule
    \multirow{2}{*}{Method} & \multicolumn{3}{c}{1\%} & \multicolumn{3}{c}{5\%} & \multicolumn{3}{c}{10\%} & \multicolumn{3}{c}{15\%} \\
    \cmidrule(lr){2-4} \cmidrule(lr){5-7}
    \cmidrule(lr){8-10} \cmidrule(lr){11-13}
          & PSNR↑ & SSIM↑ & LPIPS↓ & PSNR↑ & SSIM↑ & LPIPS↓ & PSNR↑ & SSIM↑ & LPIPS↓ & PSNR↑ & SSIM↑ & LPIPS↓ \\
    \midrule
       w/o Adaptive Selector  & 14.653 & 0.4674 & 0.5092 & 19.545 & 0.6705 & 0.3082 & 21.724 & 0.7705 & 0.2127 & 23.550 & 0.8271 & 0.1628 \\
        w/o Transform Field & 15.117 & 0.4564 & 0.6080 & 14.600 & 0.4441 & 0.6231 & 14.576 & 0.4441 & 0.6273 & 14.569 & 0.4439 & 0.6290 \\
          \midrule
     Adapt. Sel.  w/o Sparsity & 23.017 & 0.7645 & 0.2894 & 26.022 & 0.8725 & 0.1496 & 26.960 & 0.9002 & 0.1088 & 27.147 & 0.9078 & 0.0965 \\
      Adapt. Sel. w/o GI & 19.472 & 0.6325 & 0.4222 & 22.944 & 0.7702 & 0.2685 & 24.075 & 0.8155 & 0.2145 & 24.642 & 0.8360 & 0.1889 \\
        Adapt. Sel. w/o Sparsity + GI & 18.363 & 0.5528 & 0.4619 & 24.323 & 0.8336 & 0.1856 & 26.528 & 0.8919 & 0.1275 & 26.831 & 0.9013 & 0.1152 \\

        \midrule
        Full Model & \textbf{23.090} & \textbf{0.7731} & \textbf{0.2854} & \textbf{26.189} & \textbf{0.8769} & \textbf{0.1456} & \textbf{27.069} & \textbf{0.9045} & \textbf{0.1035} & \textbf{27.302} & \textbf{0.9106} & \textbf{0.0934} \\
  
    \bottomrule
    \end{tabular}
    }
    \vspace{-0.5em}
\end{table*}

\subsection{Main Results}
Experimental results for elastic inference across various ratios on Mip-NeRF360~\cite{mip360}, Zip-NeRF~\cite{barron2023zip} and T\&T~\cite{tant} are shown in Tab.~\ref{tab:sota} and Tab.~\ref{tab:sota2} respectively.
Note that different from all previous methods, which require separate training for each ratio, FlexGS only requires a single training session to enable flexible inference across different ratios, 
Visual results in Fig.~\ref{fig:main} also indicate that the proposed FlexGS offers superior efficiency and user convenience.

\paragraph{Mip-NeRF360.} The comparison results are presented in Tab.~\ref{tab:sota}. Despite utilizing less training time, our method consistently outperforms other approaches in terms of reconstruction quality across all metrics on almost all elastic ratios (10\%, 5\%, and 1\%). At lower Gaussian ratios, the performance gap becomes even more pronounced. For example, in the most challenging scenario, retaining only 1\% of the Gaussians, our method achieves a PSNR of 22.730 and an SSIM of 0.6128, outperforming LightGS* (PSNR 14.455, SSIM 0.3741) and C3DGS (PSNR 21.766, SSIM 0.5440) by a noticeable margin, indicating the significant potential of our approach for edge devices. These results underscore the robustness of FlexGS in managing varying compression levels, consistently delivering superior reconstruction quality and training efficiency.

\paragraph{T\&T.} The performance results are summarized in Tab.~\ref{tab:sota2} (left). FlexGS achieves the highest PSNR across all elastic ratios. Notably, at the challenging 1\% elastic ratio, FlexGS attains a PSNR of 23.090, substantially surpassing LightGS* (14.653 PSNR) and C3DGS (21.535 PSNR) by 8.437 and 1.555 PSNR, respectively. This performance advantage persists as the elastic ratio increases, with FlexGS reaching a PSNR of 27.802 at the 15\% ratio, again outperforming all other methods by a noticeable margin.

\paragraph{ZipNeRF.} We report the results in Tab.~\ref{tab:sota2} (right). It can be observed that FlexGS consistently delivers the best performance across all ratios. Specifically, at the 1\% ratio, FlexGS achieves a PSNR of 20.963, outperforming LightGS* (14.744 PSNR) and C3DGS (19.955 PSNR), demonstrating the robustness of FlexGS, even under extreme compression. This consistently superior performance across varying elastic ratios shows the effectiveness of FlexGS in scenes with larger scales.

\subsection{Further Empirical Study}
\paragraph{Ablation Study.}
We conduct comprehensive ablation studies to evaluate the effectiveness of each component in our framework. As shown in Tab.~\ref{tab:ablation}, we examine different configurations by disabling individual components: Adaptive Selector, Transform Field, Sparsity, and Global Importance (GI). Performance is assessed across multiple sparsity ratios (1\%, 5\%, 10\%, 15\%) using PSNR, SSIM, and LPIPS.

\begin{table}[htbp]
  \centering
  \caption{Generalized Exploration: Elastic inference performance of FlexGS under elastic ratios unseen during training, compared with other methods on ZipNeRF~\cite{barron2023zip} dataset.}
  \label{tab:gen-res}
  \vspace{-0.5em}
  \scriptsize %
  \setlength{\tabcolsep}{2.3pt}
    \begin{tabular}{c|cccccccc}
    \toprule
    \multirow{2}[2]{*}{Method} & \multicolumn{2}{c}{2\%} & \multicolumn{2}{c}{4\%} & \multicolumn{2}{c}{6\%} & \multicolumn{2}{c}{8\%} \\
      \cmidrule(lr){2-3} \cmidrule(lr){4-5}\cmidrule(lr){6-7} \cmidrule(lr){8-9}
    
          & PSNR  & SSIM  & PSNR  & SSIM  & PSNR  & SSIM  & PSNR  & SSIM \\
    \midrule
    LightGS*~\cite{fan2023lightgaussian} & 15.371  & 0.5631  & 17.582  & 0.6016  & 18.822  & 0.6714  & 19.311  & 0.6859  \\
    LightGS~\cite{fan2023lightgaussian} & 21.081  & 0.6831  & 22.448  & 0.7219  & 23.252  & 0.7467  & 23.805  & 0.7649  \\
    C3DGS~\cite{niedermayr2024compressed} & 20.524  & 0.6629  & 21.675  & 0.6936  & 22.438  & 0.7174  & 22.961  & 0.7359  \\
    EAGLES~\cite{Sharath}  & 18.493  & 0.5918  & 19.598  & 0.6039  & 20.295  & 0.6379  & 21.064  & 0.7019  \\
    Ours  & \textbf{21.636} & \textbf{0.7014} & \textbf{22.971} & \textbf{0.7406} & \textbf{24.402} & \textbf{0.7891} & \textbf{24.720} & \textbf{0.7999} \\
    \bottomrule
    \end{tabular}%
    \vspace{-1em}
\end{table}%
\begin{figure}[h]
\begin{center}
  \includegraphics[trim={7 0 7 0},clip,width=0.45\textwidth]{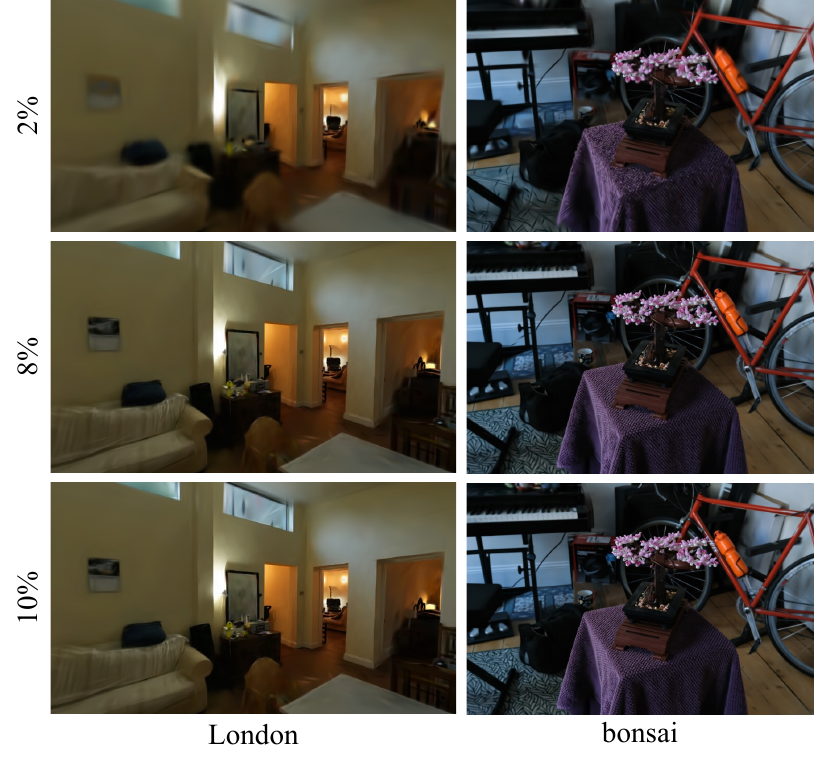}
\vspace{-1em}
\caption{Visualization of FlexGS for generalization exploration across various elastic ratios (unseen during training: $0.02$, $0.04$ and seen: $0.10$) on the ZipNeRF~\cite{barron2023zip}: \{\textit{London}\} and Mip-NeRF360~\cite{mip360}: \{\textit{bonsai}\}.
}\vspace{-2.5em}
\label{gen:res}
\end{center}
\end{figure}
Our analysis reveals that Transform Field is the most crucial component, as its removal causes severe performance degradation (PSNR drops to 14.57-15.12 across all sparsity ratios). Disabling the Adaptive Selector also significantly impacts performance (PSNR: 14.65 at 1\% sparsity), while removing GI leads to moderate degradation (PSNR: 19.47 at 1\% sparsity). The full model achieves the best performance across all metrics and sparsity ratios, with optimal results of 27.302 PSNR, 0.9106 SSIM, and 0.0934 LPIPS at 15\% sparsity, which validates the effectiveness of our complete framework design.

\paragraph{Can Elastic Ratios be Generalizable?} To evaluate whether the integration of GsNet and Gumbel-Softmax~\cite{jang2016categorical} layer, alongside the use of the dense Transform Field, enables FlexGS to generalize across varying elastic ratios, we perform elastic inference on FlexGS using ratios \{$0.02$, $0.04$, $0.06$, $0.08$\} distinct from those employed during training \{$0.01$, $0.05$, $0.10$, $0.15$\} on the Zip-NeRF~\cite{barron2023zip} dataset. Quantitative results are presented in Tab.~\ref{tab:gen-res}.
\begin{figure}[t]
\begin{center}
  \includegraphics[trim={7 0 7 0},clip,width=0.45\textwidth]{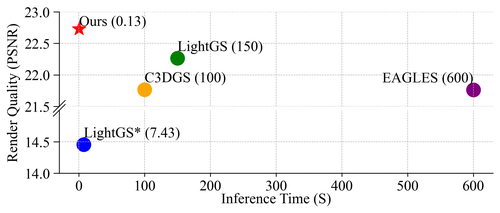}
\vspace{-1em}
\caption{
PSNR vs. inference time: Our method achieves the best balance, reducing inference time to 0.13s while maintaining high render quality (PSNR 23.0), outperforming alternatives like LightGS and EAGLES across unseen elastic ratios.
}\vspace{-2em}
\label{psnr-vs-time:res}
\end{center}
\end{figure}
As we can see, FlexGS consistently outperforms all baseline methods in elastic inference across various ratios, with improvements of at least $0.5$ in PSNR and $0.02$ in SSIM. This demonstrates its state-of-the-art performance on unseen ratios, surpassing methods with fine-tuning. Fig.~\ref{gen:res} shows that FlexGS renders outputs with novel ratios that closely match those seen during training, and the visual quality improves as the novel ratios approach the training ones. These results show the robust generalization of FlexGS across different elastic ratios, further indicating the continuity and smoothness of the Gaussian Transform Field.

\paragraph{Inference-Time Efficiency.} 
Compared to existing Gaussian compression models, our framework offers a key advantage: it achieves high-quality rendering with significantly reduced inference time, which is essential for real-time applications, especially with unseen elastic ratios. While existing methods require separate compression processes for each target ratio, our approach enables dynamic, on-the-fly compression for arbitrary ratios.
Fig.~\ref{psnr-vs-time:res} shows the trade-off between PSNR and inference time. Our method achieves the best balance, with an elastic inference time of 0.13 seconds and a high rendering quality of PSNR 23.0, outperforming other methods like EAGLES (600s) and LightGS (150s).
It also surpasses alternatives such as LightGS* (7.43s) and C3DGS (100s) in speed and quality. 
These highlight the advantage of FlexGS for maintaining rendering quality with minimal inference-time overhead.

%% file: sec/5_conclusion.tex
\section{Conclusion}
This paper presents FlexGS, a flexible 3DGS framework that enables efficient deployment across devices with varying computational constraints. Through our elastic inference method, which combines adaptive Gaussian selection and transformation modules, we achieve efficient model compression without requiring additional fine-tuning. The framework demonstrates strong adaptability by dynamically adjusting the model size according to device-specific requirements while maintaining high rendering quality. Extensive experiments on ZipNeRF, MipNeRF and Tanks\&Temples datasets validate our approach's effectiveness, showing that FlexGS successfully bridges the gap between high-quality 3D scene representation and practical deployment constraints. FlexGS provides a promising direction for making 3DGS more accessible and deployable across a broader range of applications.